# When Curiosity Signals Danger: Predicting Health Crises Through Online Medication Inquiries

Dvora Goncharok
*Department of Digital Medical Technologies*
Holon Institute of Technology

Arbel Shifman
*Department of Digital Medical Technologies*
Holon Institute of Technology

Alexander Apartsin
*School of Computer Science, Faculty of Sciences*
Holon Institute of Technology

Yehudit Aperstein
*Intelligent Systems,*
Afeka Academic College of Engineering
Tel Aviv Israel

*Abstract*—Online medical forums are a rich and underutilized source of insight into patient concerns, especially regarding medication use. Some of the many questions users pose may signal confusion, misuse, or even the early warning signs of a developing health crisis. Detecting these *critical questions* that may precede severe adverse events or life-threatening complications is vital for timely intervention and improving patient safety.

This study introduces a novel annotated dataset of medication-related questions extracted from online forums. Each entry is manually labelled for criticality based on clinical risk factors. We benchmark the performance of six traditional machine learning classifiers using TF-IDF textual representations, alongside three state-of-the-art large language model (LLM)-based classification approaches that leverage deep contextual understanding.

Our results highlight the potential of classical and modern methods to support real-time triage and alert systems in digital health spaces. The curated dataset is made publicly available to encourage further research at the intersection of patient-generated data, natural language processing, and early warning systems for critical health events.

The dataset and benchmark are available at: https://github.com/Dvora-coder/LLM-Medication-QA-Risk-Classifier-MediGuard.

## I. Introduction

The widespread availability of online health forums, Q&A platforms, and social media has transformed how patients seek and share medical information. Among these platforms, patients often turn to digital communities to voice their concerns, ask questions about medications, and report side effects or personal experiences. While these discussions provide valuable support and peer guidance, they also contain hidden signals of confusion, misuse, and potential medical crises that may go unnoticed by healthcare professionals.

Medication misuse—whether intentional or accidental—remains a pressing public health issue. It encompasses behaviors such as overdosing, using medication for unintended purposes, combining drugs in harmful ways, or deviating from prescribed regimens. Traditionally, detecting such events relies on formal channels, such as clinical reports or pharmacovigilance systems. However, these mechanisms often lag behind real-world occurrences, especially for emerging misuse patterns or new formulations. Meanwhile, user-generated content (UGC) offers a real-time, unfiltered view into patient behavior, often revealing early warning signs of risk through seemingly innocuous questions or narratives.

Recent advances in natural language processing (NLP), particularly the emergence of large language models (LLMs), offer powerful tools for understanding the nuance and intent behind user-generated health discourse. While prior research has demonstrated the potential of social media mining to detect adverse drug reactions and population-level substance use trends, less attention has been given to identifying **critical, risk-indicating patient questions**, particularly those that may signal an ongoing or impending health emergency related to medication.

This study examines the potential of automated classification methods to identify critical questions in online discussions related to medication. We begin by constructing a novel, manually annotated dataset of patient questions from medical forums, labelling them based on their potential to indicate clinical risk or misuse. We then evaluate six traditional machine learning classifiers using TF-IDF textual features, alongside three state-of-the-art LLM-based approaches that leverage contextual embeddings for classification. Our experiments highlight the comparative performance of these methods and illustrate the unique challenges posed by informal, patient-authored texts.

We publicly release our annotated dataset and accompanying benchmarks to support ongoing research. We aim to catalyze the further development of AI-driven systems capable of identifying at-risk individuals based on their online inquiries, ultimately enabling earlier interventions and improving medication safety at scale.

## II. Literature Review

Patient-generated health data (PGHD), such as wearable device outputs and social media content, has emerged as a valuable resource for inferring health conditions and behaviors. Digital phenotyping, defined as the "moment-by-moment quantification of individual-level human phenotypes using data from smartphones and other personal devices" (Jain et al., 2015), leverages PGHD to model health trajectories. For instance, studies have analyzed user-generated content from forums and Q&A sites to detect mental health conditions. De Choudhury et al. (2013) demonstrated that linguistic patterns in social media posts could predict depression, while Guntuku et al. (2017) identified markers of chronic pain in online community discussions. Chung and Basch (2015) highlighted the potential of patient forums to capture unreported symptoms in clinical settings, though biases in self-reported data remain a limitation. Ethical concerns, including privacy and data ownership, pose significant challenges in the utilization of PGHD (Martinez-Martin et al., 2018).

User-generated content (UGC) from social platforms such as forums, reviews, and health Q&A sites has become a valuable source for identifying medication misuse. These online narratives often capture real-world behaviors, such as inappropriate dosing, drug-seeking behavior, and self-medication, that are underreported in clinical settings. Recent studies have increasingly applied natural language processing (NLP) to analyze this content, uncovering both individual-level risks and population-level trends.

Early work demonstrated the feasibility of extracting misuse patterns from social media platforms, including Twitter and Reddit. For example, Sarker et al. (2020) proposed a data-centric framework for mining prescription abuse patterns, using features derived from user posts to identify potential misuse. O'Connor et al. (2020) created a Twitter-based corpus comprising over 16,000 tweets related to abuse-prone medications and introduced detailed annotation guidelines for classifying content related to misuse. Their dataset enabled the development of baseline machine learning classifiers, such as support vector machines, that demonstrated the inherent difficulty of the task due to the presence of slang, sarcasm, and subtle misuse signals.

In recent years, transformer-based large language models (LLMs) have significantly improved classification performance in this domain. Al-Garadi et al. (2021) compared several transformer architectures, including BERT and RoBERTa, to detect non-medical prescription drug use from Twitter data. Their fine-tuned models achieved substantially higher precision and recall than classical approaches, particularly for identifying nuanced or ambiguous misuse expressions. The best model reached an $F_1$-score of 0.67 for the minority (misuse) class, outperforming traditional baselines by over 20 percentage points.

Reddit has also been a popular source for studying drug misuse behaviors due to its semi-anonymous format and topic-specific communities. Garg et al. (2021) developed a method for estimating user-level fentanyl misuse risk based on Reddit posts, combining deep text embeddings with behavioral modelling. Similarly, Ge et al. (2024) introduced the Reddit-Impacts dataset, a named entity recognition resource designed to capture the clinical and social consequences of substance use.

In addition to social media, health Q&A forums offer insight into users' medication-use concerns and intentions. Kariya et al. (2023) examined user posts on Yahoo! Chiebukuro, focusing on questions about intentional overdose of over-the-counter (OTC) drugs. Their qualitative analysis revealed a pattern of risk-seeking behavior, including inquiries about euphoric effects and dosage manipulation. Although their study relied on manual coding, it highlights the potential for automated NLP tools to detect early warning signs in public forums.

Further advancements have involved mining UGC for off-label drug use and discovering emerging slang terminology. Dreyfus et al. (2021) applied NLP techniques to identify off-label drug-indication pairs from online health discussions. Holbrook et al. (2024) used Word2Vec models to discover colloquial references to opioids in Reddit posts. Such efforts enhance the coverage of vocabulary used in misuse-related discourse and support the development of more robust automated systems.

Recent reviews have underscored the promise and challenges of UGC mining for public health. Almeida et al. (2024) conducted a scoping review of opioid-related research on Reddit, raising concerns about population representativeness and affirming the value of such data for real-time surveillance. These studies demonstrate how modern NLP, particularly when powered by large language models (LLMs), can transform unstructured patient narratives into actionable insights for the early detection and monitoring of medication misuse.

Automated triage of health questions based on urgency is vital for prioritizing clinical responses. Zhang et al. (2020) fine-tuned BERT to classify consumer health questions into urgent (e.g., "chest pain") and non-urgent categories, achieving an accuracy of 89%. Earlier work by Roberts et al. (2016) used LIWC (Linguistic Inquiry and Word Count) lexicon features to flag suicidal intent in forum posts. Yang et al. (2019) developed a hierarchical model to detect questions that require professional intervention, emphasizing both syntactic and semantic cues. However, variability in layperson terminology and sarcasm in user queries pose challenges (Benamara et al., 2018).

Machine learning models are increasingly used to triage high-risk text in clinical workflows. Jiang et al. (2021) compared CNN and LSTM models for classifying urgent patient messages in telehealth platforms, finding that hybrid architectures outperformed traditional methods. Klein et al. (2020) demonstrated that ensemble models reduced false negatives in detecting suicidal ideation in EHR notes. Chen et al. (2019) highlighted the role of active learning in maintaining classifier performance with limited labelled data. Key challenges include minimizing false positives in critical alerts and ensuring model interpretability for clinicians (Wallace et al., 2020).

## III. METHODOLOGY

### A. Dataset

We annotated a sample from the MedInfo2019-QA-Medications dataset, which is closely associated with the MEDIQA 2019 Shared Task, which aims to advance medical question answering (QA) systems through tasks such as natural language inference (NLI) and recognizing question entailment (RQE) (Ben Abacha et al., 2019). While the specific name "MedInfo2019-QA-Medications" is not explicitly used in the published literature, the dataset aligns with the MedQuAD (Medical Question Answering Dataset) developed by Abacha, Shivade, and Demner-Fushman (2019).

| Question | Risk Level |
|---|---|
| Is it safe to take ibuprofen while on blood thinners? | Critical |
| Can I split my metformin pill to reduce the dose? | General |
| What happens if I accidentally take 50mg of lisinopril twice? | Critical |
| Is it okay to drink grapefruit juice with my medication? | General |

Table 1: Examples of critical and general user queries

This dataset comprises 47,457 medical question-answer pairs from 12 authoritative U.S. National Institutes of Health (NIH) websites, including MedlinePlus and Cancer.gov (Abacha et al., 2019). We have labelled 650 examples with a binary risk level label as "general" or "critical." The data is highly imbalanced, with only around 100 questions labelled "critical".

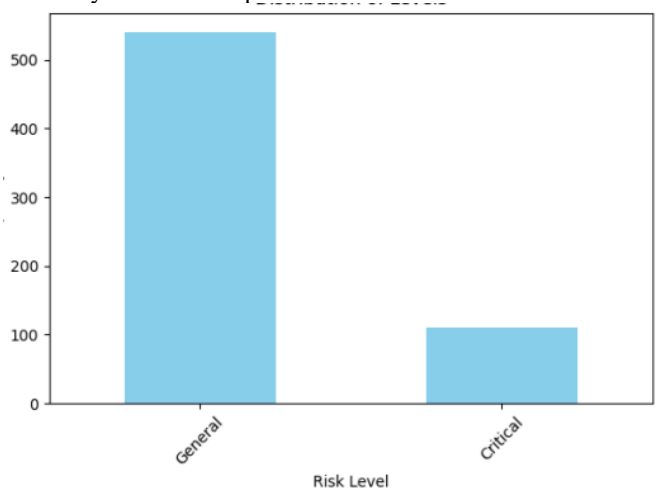

*Figure 1: Class distribution in the generated dataset*

### B. Classification with classical machine learning models

First, text preprocessing cleans the data by removing irrelevant characters, stop words, and formatting issues. Next, tokenization splits the cleaned text into individual words, or tokens, which serve as the basic units for analysis. Then, TF-IDF vectorization converts these tokens into numerical vectors, capturing the relative importance of each word within the dataset.

Additionally, feature engineering is applied to enhance the data, including the creation of a Critical Similarity feature to enrich the analysis further. The Critical Similarity feature measures the closest cosine distance between the TF-IDF vector of a given text sample and the TF-IDF vector of a critical training question.

As part of our preprocessing pipeline for traditional machine learning models, we apply dimensionality reduction to the TF-IDF representations of the text data. While sparse and high-dimensional, the original TF-IDF vectors capture word frequency and importance, but they often lead to noisy or computationally expensive representations. To address this, we use Singular Value Decomposition (SVD), a matrix factorization technique, to reduce the dimensionality of the TF-IDF feature space. Specifically, we perform truncated SVD (also known as Latent Semantic Analysis, or LSA), which identifies a lower-dimensional representation of the data by projecting it onto the directions of maximum variance in the original term-document matrix.

### C. Classification with large language models

We implemented and compared three LLM-based classification strategies to evaluate the effectiveness of large language models (LLMs) in detecting critical health-related questions. These approaches leverage a deep contextual understanding, allowing models to interpret user intent, implied risk, and subtle cues that are often missed by classical methods. Below are the three most promising strategies we employed:

We used the standard BERT-based model and **fine-tuned** it on our annotated dataset of medication-related questions. This approach treated classification as a supervised task, where each input question was fed through the model and a SoftMax layer predicted the probability of the "critical" vs "non-critical" label. Despite being a general-purpose model, fine-tuning BERT on task-specific data allowed it to adapt to informal medical language, common concerns, and patient phrasing patterns found in user-generated content. It served as a strong baseline and outperformed classical models across all metrics.

To incorporate medical domain knowledge, we fine-tuned **BioBERT**, a version of BERT pre-trained on large-scale biomedical corpora, including PubMed abstracts and PMC full-text articles. While BioBERT was not trained initially on informal or patient-authored texts, its rich understanding of medical terminology and drug-condition relationships made it particularly effective in identifying questions involving unsafe drug use, harmful interactions, or medically urgent symptoms. BioBERT demonstrated higher precision in identifying clinically relevant risk patterns, especially in questions that mentioned specific drug names, adverse effects, or comorbid conditions.

BlueBERT differs from BioBERT in both its training sources and clinical orientation. Whereas BioBERT is trained solely on biomedical literature such as PubMed abstracts and PMC full texts, BlueBERT combines this biomedical corpus with de-identified clinical notes from the MIMIC-III database. This additional exposure to hospital documentation equips BlueBERT with a firmer grasp of real-world clinical language abbreviations, shorthand, and context-rich patient narratives—which are rare in research papers. As a result, it often outperforms BioBERT on tasks that require understanding electronic health records or bedside terminology, such as detecting treatment complications, capturing nuanced temporal events, or mapping symptoms to procedures. To assess these advantages in practice, we included a BlueBERT-based classifier in our benchmark, allowing direct comparison with BioBERT and the GPT model.

We explored **prompt-based few-shot** classification using an instruction-tuned GPT model as a third strategy. Instead of fine-tuning, we used natural language prompts that framed the task (e.g., *"Is this question medically urgent or potentially harmful? Respond with Yes or No"*). Even with limited supervision, the model could generalize well by providing a few labelled examples in the prompt. This method required no parameter updates and was particularly effective in recognizing nuanced linguistic signals, such as desperation, uncertainty, and misused intent. While more sensitive to prompt phrasing, GPT-style models excelled at borderline cases where others struggled.

## IV. RESULTS

We split the annotated dataset into an 80/20 train-test split to evaluate model performance, ensuring a balanced representation of critical and non-critical questions in both subsets. We applied 5-fold cross-validation on the training set for the classical machine learning models to tune hyperparameters and assess generalization performance before final testing.

| Type | Method | Accuracy | F1 |
|---|---|---|---|
| Classical ML Model | SVM (baseline) | 0.84 | 0.80 |
| | Logistic Regression | 0.76 | 0.77 |
| | Gradient Boosting | 0.79 | 0.77 |
| | Random Forest | 0.68 | 0.70 |
| | SGD Logistic (L2) | 0.79 | 0.79 |
| Large Language Models | BioBERT | **0.92** | **0.90** |
| | BlueBERT | 0.91 | 0.90 |
| | GPT-4.1 | 0.87 | 0.85 |

**Table 2**: Summary of the model evaluation results

These classical models operated on TF-IDF representations of the text, further processed through dimensionality reduction as described above. In contrast, the large language model (LLM)-based approaches were applied directly to the raw or lightly pre-processed text.

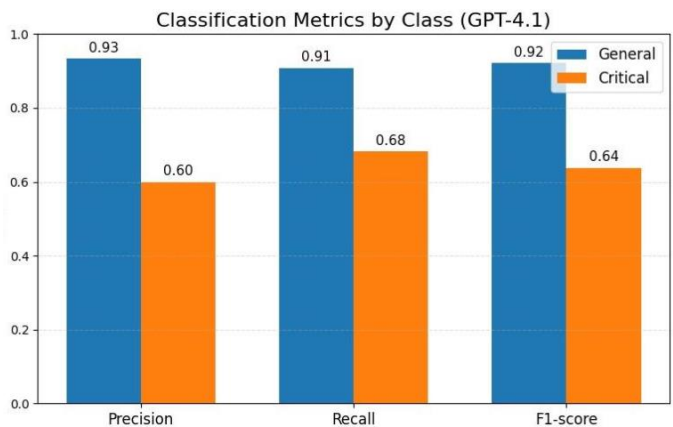

*Figure 2: Performance Metrics of GPT-4.1-based classifier*

Where appropriate, data augmentation and prompt engineering strategies were employed to enhance generalization for large language models (LLMs). This setup allowed us to compare traditional feature-based classifiers against end-to-end language model approaches on a common evaluation framework.

## V. CONCLUSIONS AND FUTURE RESEARCH

In this study, we investigated natural language processing techniques, ranging from traditional machine learning to large language models, to detect critical health risks in patient-generated medication questions. By curating and manually annotating a dataset of real-world questions from health forums, we created a focused resource that captures user concerns potentially indicative of medication misuse, confusion, or adverse effects. Our evaluation of six classical classifiers and three LLM-based approaches demonstrated that transformer-based models significantly outperform traditional baselines in identifying high-risk inquiries, especially those fine-tuned on biomedical text.

A key contribution of this work is the release of a publicly available annotated dataset, designed to support future research on the automatic detection of medically urgent or high-risk user queries. In addition to establishing benchmark results, we have highlighted the unique challenges posed by informal, patient-authored content, such as ambiguous language, a lack of clinical terminology, and emotionally charged phrasing, which necessitate models with a deeper understanding of context.

Looking forward, several directions warrant further exploration. First, our current binary classification approach (critical vs. non-critical) can be extended to more fine-grained risk stratification, capturing varying degrees of urgency or severity. A multi-tiered risk labelling scheme could enable more nuanced triage and prioritization of patient concerns. Second, incorporating multimodal signals such as user history, drug metadata, or symptom ontologies may improve model performance and explainability.